\documentclass[10pt,twocolumn,letterpaper]{article}

 \usepackage[pagenumbers]{cvpr}              
\definecolor{cvprblue}{rgb}{0.21,0.49,0.74}
\usepackage[pagebackref,breaklinks,colorlinks,allcolors=cvprblue]{hyperref}

\def\paperID{15104} 
\def\confName{CVPR}
\def\confYear{2026}

\title{ViLaCD-R1: A Vision-Language Framework for Semantic Change Detection in Remote Sensing}
\usepackage{float}

\author{
Xingwei Ma\\
Fudan University\\
Shanghai\\
{\tt\small 24210720233@m.fudan.edu.cn}
\and
Shiyang Feng\\
Shanghai Artificial Intelligence Laboratory\\
Shanghai\\
{\tt\small fengshiyang@pjlab.org.cn}
\and
Bo Zhang\\
Shanghai Artificial Intelligence Laboratory\\
Shanghai\\
{\tt\small bo.zhangzx@gmail.com}
\and
Bin Wang\\
Fudan University\\
Shanghai\\
{\tt\small wangbin@fudan.edu.cn}
}

\begin{document}
\maketitle
\begin{abstract}
Remote sensing change detection (RSCD), a complex multi-image inference task, traditionally uses pixel-based operators or encoder-decoder networks that inadequately capture high-level semantics and are vulnerable to non-semantic perturbations. Although recent multimodal and vision-language model (VLM)-based approaches enhance semantic understanding of change regions by incorporating textual descriptions, they still suffer from challenges such as inaccurate spatial localization, imprecise pixel-level boundary delineation, and limited interpretability.
To address these issues, we propose ViLaCD-R1, a two-stage framework comprising a Multi-Image Reasoner (MIR) and a Mask-Guided Decoder (MGD). Specifically, the VLM is trained through supervised fine-tuning (SFT) and reinforcement learning (RL) on block-level dual-temporal inference tasks, taking dual-temporal image patches as input and outputting a coarse change mask. Then, the decoder integrates dual-temporal image features with this coarse mask to predict a precise binary change map.
Comprehensive evaluations on multiple RSCD benchmarks demonstrate that ViLaCD-R1 substantially improves true semantic change recognition and localization, robustly suppresses non-semantic variations, and achieves state-of-the-art accuracy in complex real-world scenarios.
\end{abstract}    
\section{Introduction}
\label{sec:intro}
\begin{figure}[t]
\centering
\begin{subfigure}{0.33\linewidth}
    \centering
    \includegraphics[height=4cm]{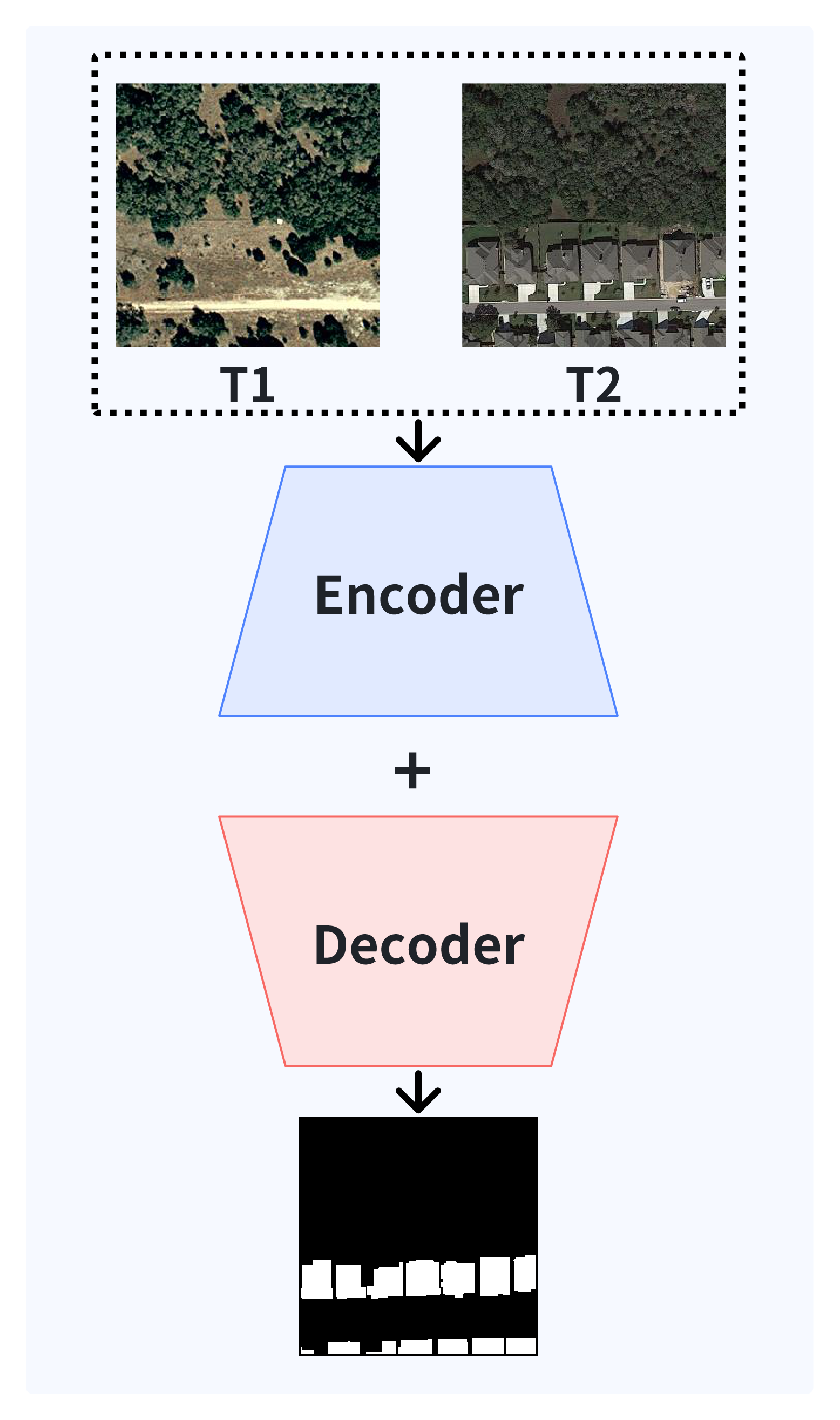}
    \caption{}
    \label{fig:sub1}
\end{subfigure}%
\hspace{0.01\linewidth}%
\begin{subfigure}{0.66\linewidth}
    \centering
    \includegraphics[height=4cm]{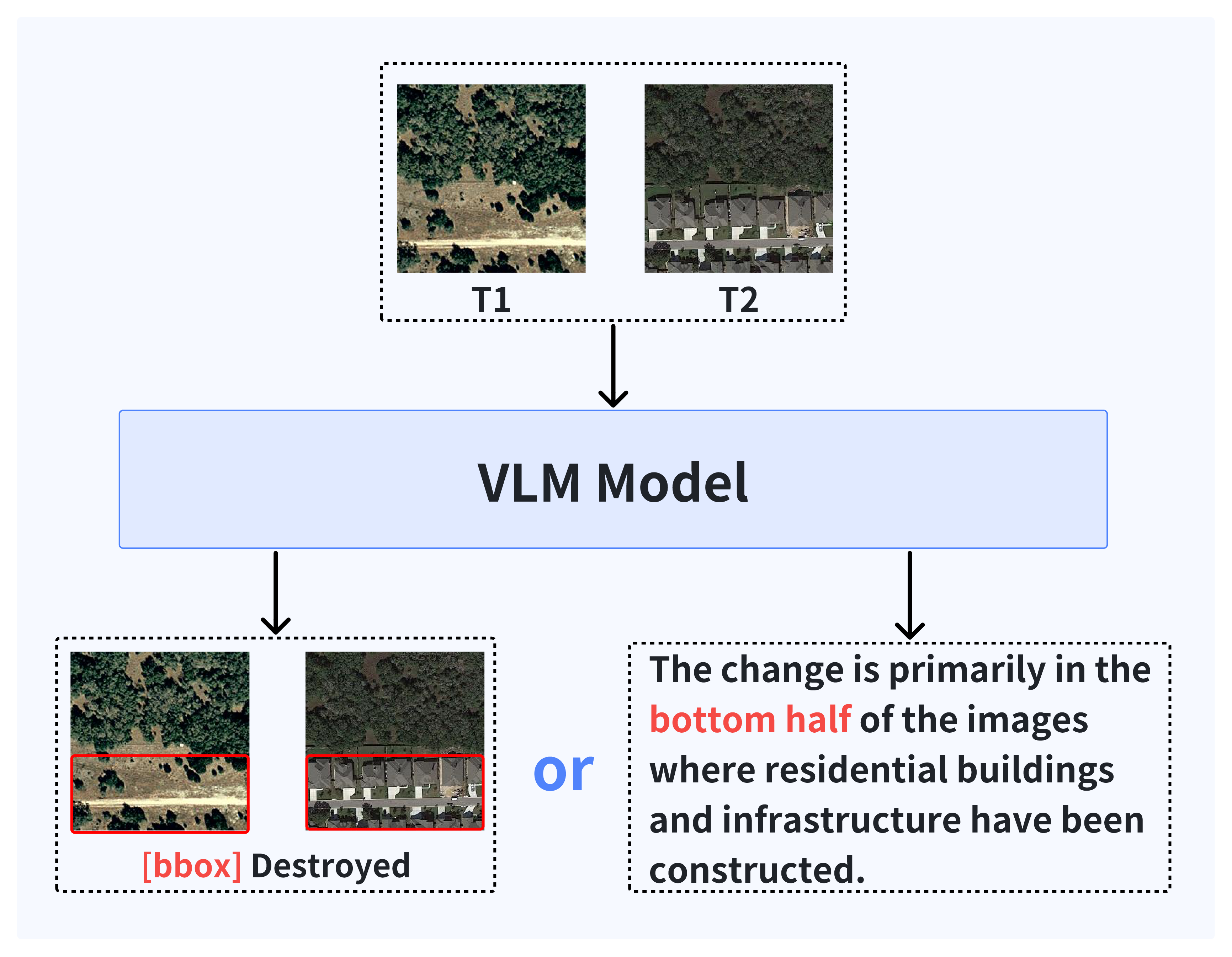}
    \caption{}
    \label{fig:sub2}
\end{subfigure}

\vspace{0.1cm}  

\begin{subfigure}{\linewidth}
    \centering
    \includegraphics[width=\linewidth]{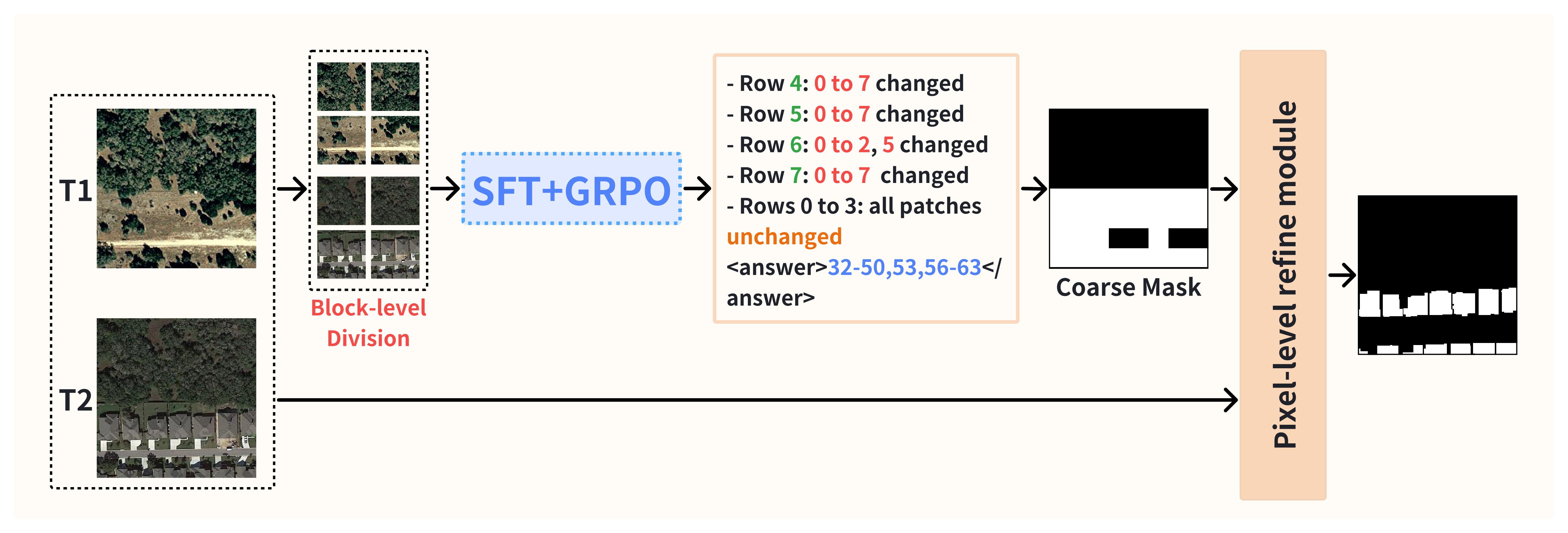}
    \caption{}
    \label{fig:sub3}
\end{subfigure}

\caption{(a) illustrates the feature-based paradigm for remote sensing change detection; (b) shows the VLM-based paradigm; and (c) presents our proposed ViLaCD-R1 architecture, which provides a complete pipeline from coarse region prompting to fine-grained pixel-level segmentation.}
\label{fig:main}
\end{figure}
 
Remote sensing change detection (RSCD) identifies land cover variations by analyzing multi-temporal remote sensing images (RSIs), playing a vital role in urban monitoring, disaster assessment, and ecological analysis~\cite{zheng2024single, wei2024spatio, DLCD, feng2024psd}. With the rapid growth of high-resolution RSIs, achieving accurate, semantically consistent, and generalizable change detection under complex scenarios has become a central challenge.

With the development of deep learning, RSCD has gradually shifted from traditional hand-crafted feature engineering toward end-to-end feature learning paradigms. Feature-based methods constitute the mainstream of deep learning–driven RSCD, relying on CNNs, Transformers, and hybrid architectures to extract multi-scale, multi-level spatial–spectral representations. Early CNN-based approaches, with representative works such as FCSN~\cite{FCSN} and DASNet~\cite{dasnet}, focus on hierarchical spatial features, while later Transformer-based models, including STANet~\cite{sta} and ChangeFormer~\cite{changeformer}, strengthen global context modeling through self-attention.
However, feature-based methods still struggle to capture higher-level semantic change cues due to limited semantic abstraction and difficulty in modeling cross-temporal reasoning, often leading to false alarms in texture-dominated regions and missed detections for subtle man-made changes.

Motivated by the success of large multimodal model ~\cite{chen2024internvl, bai2025qwen2}, Vision–Language Model (VLM)-based methods have emerged as a new paradigm for RSCD, leveraging pretrained visual priors to infer change categories or describe scene transitions. These approaches utilize the rich world knowledge and cross-modal alignment learned from large-scale corpora to enhance semantic discrimination, enabling more robust identification of meaningful land-cover transitions~\cite{VLP, DLCD}. VLM-based methods can mitigate the semantic limitations of feature-based paradigms by providing richer contextual cues and stronger invariance across imaging conditions ~\cite{deng2025changechat,li2024semicd}. Nevertheless, VLM generally lack fine-grained spatial localization ability, as their coarse visual tokenization and global attention prevent accurate boundary reconstruction and pixel-level consistency.

Overall, feature-based and VLM-based methods present complementary strengths: feature-based models offer strong spatial sensitivity but weak semantics, whereas VLM-based models provide strong semantic reasoning but weak spatial localization. This fundamental gap is illustrated in Fig.~\ref{fig:main}, highlighting the need for a framework that simultaneously achieves semantic reasoning and precise spatial change detection.

To address these issues, we propose \textbf{ViLaCD-R1}, a two-stage framework comprising a fine-tuned Vision–Language Model as the \textbf{Multi-Image Reasoner (MIR)} and a \textbf{multi-layer multi-scale Mask-Guided Decoder (MGD)}. Specifically, the MIR is trained through supervised fine-tuning (SFT) and reinforcement learning (RL) on block-level dual-temporal inference tasks, taking dual-temporal RSI patches as input and outputting a coarse change mask. Subsequently, the MGD integrates deep dual-temporal image features with this coarse mask to predict a precise binary change map.
Moreover, during the RL stage, ViLaCD-R1 introduces a task-adaptive reward function based on Group Policy Optimization (GRPO) to explicitly enhance cross-temporal consistency and improve the discrimination of authentic changes. Extensive experiments on multiple RSCD datasets demonstrate superior robustness, interpretability, and cross-scene generalization.

\noindent

The main contributions are summarized as follows:

\begin{itemize}[leftmargin=10pt,itemsep=2pt,topsep=2pt]
\item \textbf{Unified Dual-Stage Architecture:} ViLaCD-R1 is designed as a unified two-stage framework that couples semantic-level reasoning from a vision–language model with pixel-level spatial decoding. This design resolves the long-standing challenge of jointly achieving high-level semantic understanding and precise spatial localization in remote sensing change detection.
\item \textbf{Multi-Image Reasoner:} We develop a fine-tuned VLM as the Multi-Image Reasoner, trained through SFT and GRPO-based RL on block-level bi-temporal inference tasks. The MIR effectively captures semantic cross-temporal relationships and produces coarse yet semantically reliable change masks that guide downstream reconstruction.

\item \textbf{Mask-Guided Decoder:} We design a multi-layer, multi-scale Mask-Guided Decoder that integrates dual-temporal deep features with MIR-generated masks to perform adaptive multi-granularity distillation. This decoder enhances fine-grained spatial localization and boundary reconstruction, yielding more accurate and interpretable change maps.
\end{itemize}


\section{Related Work}
\label{sec:relative}

\subsection{Feature-based Change Detection Methods}

Recent advances in remote sensing change detection have largely centered on extracting and fusing features from bi‑temporal images, followed by fine‑grained decoding~\cite{feng2025df}. Deep learning has significantly strengthened this pipeline. CNN-based methods learn hierarchical representations directly from data and enable end‑to‑end pixel‑level prediction, surpassing traditional hand‑crafted features~\cite{ASDL}. Siamese architectures are widely used, where two shared encoders process the two timestamps and their differences are modeled afterward. For example, SNUNet‑CD \cite{SNUNet-CD} uses dense connections and multi‑scale fusion to enhance change discrimination. Attention mechanisms are often added to highlight subtle or small‑scale changes.

More recently, Transformer-based approaches have been introduced to capture long‑range dependencies through self‑attention, which is particularly useful for high‑resolution imagery and complex contexts~\cite{ATBM}. Models such as SwinSUNet~\cite{SwinSUNet}, ChangeCLIP~\cite{ChangeCLIP}, and TransUNetCD~\cite{TransUNet-CD} demonstrate strong capabilities in multi‑scale modeling and complex‑scene recognition. Compared with CNNs, Transformers better capture cross‑temporal global context, but often require additional local‑detail enhancement or specialized decoders to recover precise boundaries and remain efficient on large images.

\subsection{VLM-based Change Detection Methods}
VLM, such as CLIP, BLIP, and Qwen-VL, have demonstrated strong potential in cross‑modal semantic alignment and reasoning. By jointly leveraging visual and language representations, VLM enable semantic understanding, compositional reasoning, and contextual interpretation, providing a new direction for semantic‑level change detection~\cite{EFMRSI}.

Recent VLM‑driven methods attempt to input multi‑temporal remote sensing imagery into large-scale pretrained models to directly generate change categories, semantic descriptions, or coarse localization maps. For example, TEOChat~\cite{TEOChat}, Change-Agent~\cite{Change-Agent}, and ChangeChat~\cite{ChangeChat} integrate VLM with image‑to‑image contrastive tasks, using natural language prompts to produce change‑region bounding maps that guide change detection. CKCD~\cite{CKCD} further incorporates change knowledge into semantic labels to enhance clarity and consistency, improving semantic‑level guidance for change detection.

These methods can more accurately distinguish semantic changes (such as building construction or demolition, or urban green‑space variation) from non‑essential disturbances (such as seasonal vegetation fluctuation or shadows), while greatly reducing reliance on pixel‑level annotations. However, they remain limited in fine‑grained pixel‑level localization and small-object detection. Their coarse masks are insufficient for high‑resolution change detection tasks requiring precise boundary delineation, often making downstream decoders or feature‑refinement modules necessary to achieve pixel‑level predictions.

\subsection{VLM for multi-image reasoning}
Recent studies have shown that integrating RL into VLM enhances multi‑modal reasoning and supports more complex logical inference~\cite{feng2025dsf2}. For example, RARL applies rule‑based reward augmentation under the GRPO framework to improve reasoning ability in medical VLM, achieving notable gains on benchmarks such as VQA‑RAD while maintaining efficiency through synthetic data~\cite{RARL}.

Similarly, Self‑Rewarding VLM decompose reasoning into visual description and language deduction, effectively reducing hallucinations and achieving stronger performance on ChartQA compared with conventional baselines~\cite{SRVLM}. In addition, Visual‑RFT~\cite{Visual-RFT} and VLM‑R1~\cite{VLM-R1} extend RL techniques to image‑perception tasks, while Geo‑R1~\cite{Geo-R1} introduces a reasoning‑centric Reinforcement Fine‑Tuning (RFT) paradigm for few‑shot geospatial reference tasks, substantially surpassing standard SFT methods.

Together, these approaches demonstrate that RL‑driven VLM reasoning is highly effective for structured multi‑modal tasks and provides a strong foundation for advancing semantic‑visual alignment.


\section{Method}

In this section, we first define the change detection task and its core challenges, followed by an overview of the proposed \textbf{ViLaCD-R1} framework, which integrates semantic block construction, structured representation learning, and reasoning-driven coarse change localization. We then introduce the coarse-to-fine decoder module guided by the coarse semantic mask, which balances semantic interpretability and spatial precision. Finally, we summarize implementation details and experimental setup.

\subsection{Preliminary}

\subsubsection{Problem Definition}

Given two temporally aligned remote sensing images $I_{t_1}$ and $I_{t_2}$, the objective is to predict a binary change map 
\[
M \in \{0, 1\}^{H \times W},
\]
where $M_{ij} = 1$ indicates a semantic change occurring at position $(i,j)$ between the two timestamps, and $M_{ij} = 0$ otherwise.  
Traditional approaches typically formulate this as a pixel-wise classification problem by learning a mapping function
\[
f_\theta : (I_{t_1}, I_{t_2}) \rightarrow M,
\]
which directly predicts pixel-level changes. However, such methods mainly rely on low-level spectral or textural differences, making them sensitive to illumination, seasonal, and sensor variations, and limiting their ability to capture high-level semantic changes~\cite{ASDL}.

\subsubsection{RL for fine-tuning VLM}
RL is widely used for post‑training large models such as LLMs and VLM. In this work, we adopt GRPO, which eliminates the need for a value model and normalizes rewards within groups of outputs generated from the same prompt. This design yields more stable advantage estimates and reduces computational overhead, making it well suited for 
sequence‑level optimization.
Given a prompt $s \sim p_{\text{prompt}}$, the policy $\pi_{\theta_{\text{old}}}$ generates $G$ samples $\{o_i\}$ with rewards $r_i$. GRPO computes
\[
\hat{A}_i = \frac{r_i - \mu}{\max(\sigma, \varepsilon)},
\]
and optimizes a PPO-style clipped objective:
\[
J_{\text{GRPO}}(\theta)
= \mathbb{E}\!\left[
\frac{1}{G}\sum_{i=1}^G 
\text{clip}\!\big(r_i(\theta)\hat{A}_i\big)
\right].
\]

Although originally introduced for preference align-
ment, GRPO naturally extends to objective tasks with ver-
ifiable ground truth, such as classification, matching, nu-
merical prediction, and change detection. In these settings,
rewards can be directly defined based on consistency with
ground truth, eliminating the need for a learned preference
model and enabling broad applicability across language and
vision domains.

\subsection{Architecture}
\begin{figure*}[t]
    \centering
    \includegraphics[width=1.0\linewidth]{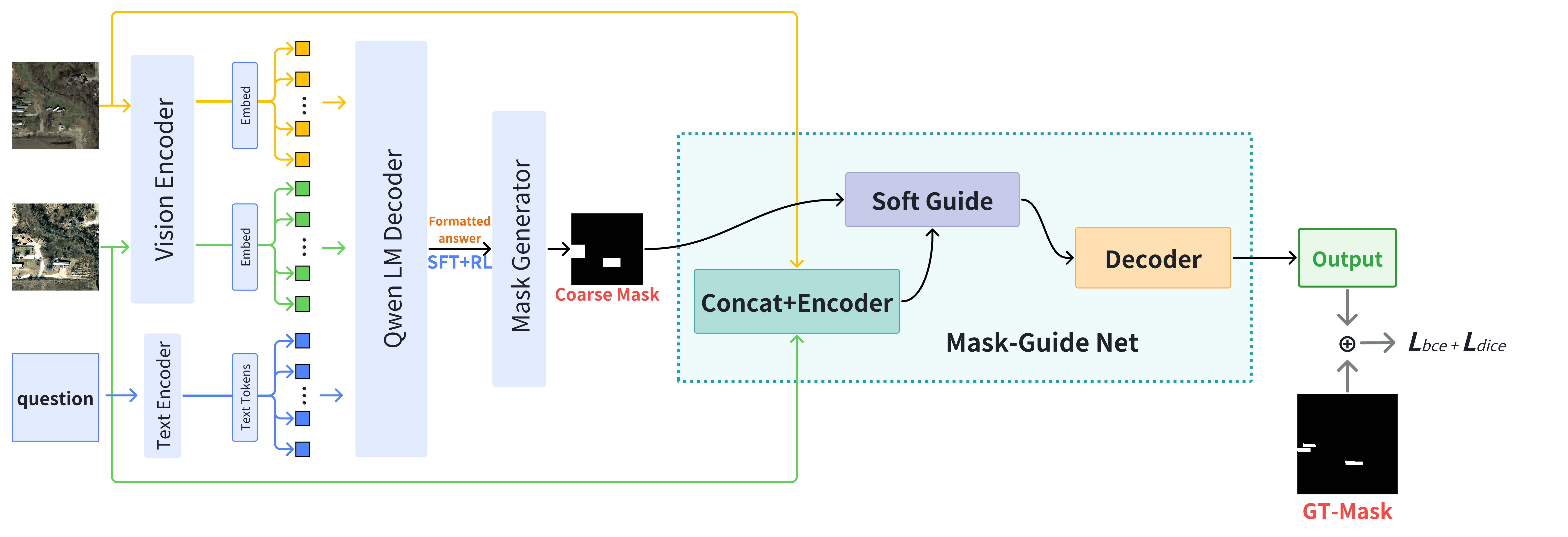}
    \caption{The overall architecture of ViLaCD-R1. The MIR module first generates coarse change masks for the bi-temporal images. These masks are then used to guide MGD-Net, enabling fine-grained, pixel-level boundary refinement.}
    \label{fig:architecture}
\end{figure*}
\subsubsection{Overview}
In ViLaCD-R1, we reformulate change detection as a two-stage hierarchical reasoning process, as illustrated in Fig.~\ref{fig:architecture}. Specifically, the first stage performs semantic-level coarse detection, in which a VLM infers patch-level semantic changes, thereby enabling cross-modal reasoning and fostering a comprehensive global understanding of the scene. Then, the second stage  conducts mask-level fine detection, reconstructing high-resolution change masks under the guidance of the coarse predictions, thus achieving precise spatial localization. Consequently, this two-stage design establishes collaborative reasoning across semantic and spatial dimensions, ultimately producing change maps that are both semantically interpretable and spatially accurate.
\subsubsection{MIR Training}

Traditional multimodal tasks in remote sensing change detection typically provide only coarse semantic descriptions, which are insufficient to support fine-grained mask segmentation requirements. To fully leverage the powerful textual reasoning capabilities of VLM, we reformulate the change detection task as a structured semantic change prediction problem. Inspired by the Text4Seg~\cite{text4seg} approach, we partition the input bi-temporal images into $8 \times 8$ equally-sized patches and index them in row-major order (0 to 63). Each index corresponds to a potential semantic change region of size $[H/8, W/8]$ in the original image.

Within this framework, a binary decision is performed for each image patch: if a semantic change is present, the patch is labeled as changed; otherwise, it is labeled as unchanged. The indices of all patches identified as changed are formatted into a structured output in row order. This structured output mechanism guides the model to learn patch-wise semantic comparison, effectively enhancing the model's interpretability and reasoning consistency. To optimize computational efficiency, adjacent changed indices are merged during the output stage to reduce token usage.

\begin{figure}[h]
    \centering
    \includegraphics[width=1.0\linewidth]{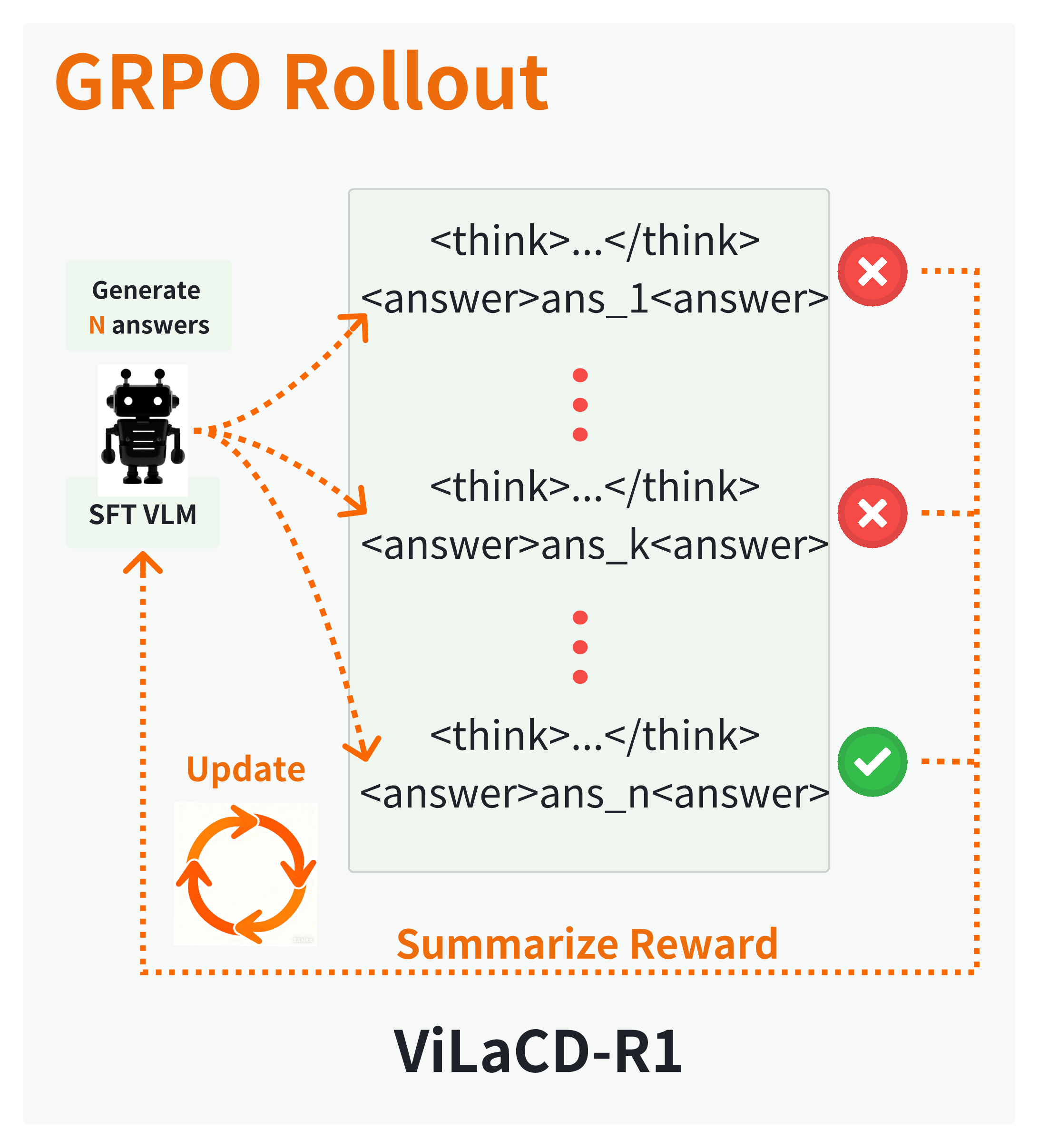}
    \caption{GRPO training workflow}
    \label{fig:GRPO}
\end{figure}

Based on this structured output format, the training of the vision-language model is conducted in two stages:
\begin{itemize}
    \item \textbf{SFT Stage:} Utilizes formatted output data to adapt the VLM to the visual-language alignment requirements specific to remote sensing imagery.
    \item \textbf{RL Stage:} Employs task-related reward functions to guide the model in accurately identifying genuine semantic changes while suppressing irrelevant discrepancies (the specific mechanism is illustrated in Figure~\ref{fig:GRPO}).
\end{itemize}

Finally, the index sequence output by the model is converted into a coarse semantic mask through spatial mapping, providing a foundation for subsequent fine segmentation.

\subsubsection{MGD Structure}

To achieve pixel-level precise localization, we propose a guided change detection decoder termed MGD, based on the U-Net architecture. This module employs an encoder-decoder structure and innovatively integrates a local window Transformer with a soft guidance mechanism to improve the accuracy of bi-temporal remote sensing image change detection.

In the encoder stage, we adopt a multi-layer convolutional downsampling structure. Through successive spatial reduction and channel expansion, it progressively extracts multi-scale hierarchical feature representations from the input images, thereby constructing a feature pyramid that spans from local details to global semantics.

To balance computational efficiency with the trade-off between local and global contextual information, we introduce a local window Transformer module. This module partitions the feature maps into adaptively-sized local windows and performs self-attention computation within each window, effectively capturing long-range dependencies. This design aids in distinguishing genuine land-cover changes from pseudo-changes caused by illumination variations, seasonal shifts, and other confounding factors, while simultaneously accommodating change regions of varying scales.

To further enhance the model's performance, we design a soft guidance mechanism to better utilize the prior coarse masks generated by the MIR module. This mechanism enhances features exclusively in regions suspected of change as indicated by the mask, while preserving the original feature strength in background areas. By employing a pyramidal multi-resolution mask generation strategy, we ensure precise spatial alignment of the mask across different decoding stages. Moreover, we introduce a learnable guidance strength coefficient $\alpha$, enabling the model to adaptively regulate the contribution weight of the prior mask during feature enhancement, thereby achieving a balance between data-driven learning and prior knowledge integration.

\subsection{Reward Design}

To encourage reasoning beyond the predefined framework, the reward function includes both a format reward and a task-specific answer reward.

\noindent\textbf{Format Reward:} To ensure reliable evaluation of model outputs, we require explicitly defined output formatting. The reasoning must be enclosed in \texttt{<think>...\texttt{</think>}}, and the final answer in \texttt{<answer>...\texttt{</answer>}} as a merged index string such as $i\!-\!j,k,\dots,n$. The format reward is defined as:  
\begin{equation}
R_{\text{format}} = 
\begin{cases}
1, & \text{if output follows the expected format} \\
0, & \text{otherwise.}
\end{cases}
\end{equation}

\noindent\textbf{Prediction Reward:} For the coarse-grained change mask prediction task, the VLM outputs a list of indices of all changed blocks, denoted as $L_{\text{pred}}$, which is then passed to a mask generator to produce the coarse-grained mask. We define the precision-recall reward as:

\begin{equation}
R_{\text{acc}} = (1-\beta) \cdot \text{Precision} + \beta \cdot \text{Recall},
\end{equation}
Since the prediction mainly focuses on covering the change regions in the image, we set $\beta = 0.7$ to emphasize recall and ensure sufficient coverage of changed areas.

\noindent\textbf{Recall Bonus:} To further encourage higher recall, we introduce a recall bonus $R_{\text{bonus}}$ defined as:

\begin{equation}
R_{\text{bonus}} =
\begin{cases}
0.7, & \text{if } \text{Recall} > 0.7, \\
0.9, & \text{if } \text{Recall} > 0.9, \\
1.0, & \text{if } \text{Recall} = 1.0.
\end{cases}
\end{equation}

\subsection{Decoder Design}

After obtaining the coarse-grained mask, although it provides an interpretable overview of the change regions, a fine-grained decoder is still required to generate precise pixel-level change masks. Inspired by MaskCD~\cite{MaskCD}, we design \textbf{MGD-Net}, a decoder module for binary change detection (BCD) in remote sensing images, whose core objective is to combine the coarse-grained mask with bi-temporal image features to achieve pixel-level fine-grained change prediction. The network architecture is illustrated in Fig.~\ref{fig:architecture}.

The decoder follows an encoder-bottleneck-decoder structure, incorporating local window mechanisms to enhance global semantic modeling capabilities~\cite{ELGC-Net,TransY-Net}. Within the decoder, we introduce mask-guided feature enhancement, which significantly improves fine-grained prediction for small targets and boundaries. Furthermore, during the decoding stage, we integrate local window self-attention to model global context in high-resolution imagery, effectively capturing long-range dependencies and addressing the limitations of traditional CNNs in semantic change understanding.

Additionally, we propose a soft guidance strategy that weights features within mask regions while preserving information in non-mask areas, improving the model's robustness. With this design, MGD-Net simultaneously achieves semantic-level prior guidance and pixel-level boundary refinement, providing an efficient solution for high-resolution remote sensing change detection that balances both accuracy and semantic understanding.

\begin{table*}[h]
\centering
\caption{Performance comparison of different change detection models on LEVIR-CD dataset.}
\label{tab:levir}
\vspace{-0.3cm}
\normalsize  
\setlength{\tabcolsep}{12pt}  
\begin{tabular}{lccccc}
\toprule
\textbf{Method} & \textbf{Precision (\%)} & \textbf{Recall (\%)} & \textbf{F1 (\%)} & \textbf{IoU (\%)} & \textbf{OA (\%)} \\
\midrule
FC-Siam-conc~\cite{ASTABM}   & 88.79 & 86.57 & 87.67 & 78.04 & 98.75 \\
FC-Siam-diff~\cite{FC-Siam-diff}   & 89.25 & 82.62 & 85.81 & 75.14 & 98.46 \\
STANet~\cite{sta}         & 90.68 & 87.70 & 89.17 & 80.45 & 98.91 \\
SNUNet~\cite{SNUNet-CD}         & 91.25 & 85.55 & 88.30 & 79.06 & 98.85 \\
HANet~\cite{HANet}          & 91.21 & 89.36 & 90.28 & 82.27 & 99.02 \\
Changeformer~\cite{ATBSN}   & 91.85 & 87.88 & 89.82 & 81.52 & 98.99 \\
SwinSUNet~\cite{SwinSUNet}      & 90.76 & 86.92 & 88.80 & 79.85 & 98.88 \\
BIT~\cite{BIT}            & 91.74 & 88.25 & 89.96 & 81.76 & 99.00 \\
ConvTransNet~\cite{ConvTransNet}   & 92.64 & 88.58 & 90.56 & 82.75 & 99.06 \\
WNet~\cite{WNet}           & 91.16 & 90.18 & 90.67 & 82.93 & 99.06 \\
HMCNet~\cite{HMCNet}         & 91.68 & 89.82 & 90.74 & 83.05 & 99.07 \\
\midrule
\textbf{ViLaCD-R1(Ours)} & \textbf{92.76} & \textbf{90.84} & 87.70 & \textbf{84.33} & 98.99 \\
\bottomrule
\end{tabular}
\vspace{-0.3cm}
\end{table*}

\subsection{Loss Function for Binary Change Detection}

To address the severe class imbalance in remote sensing binary change detection, where foreground (changed) pixels typically occupy a small fraction of the image, we design a composite loss function that combines weighted binary cross-entropy (BCE) and Dice loss.

Formally, let $P \in [0,1]^{B \times 1 \times H \times W}$ denote the predicted probability map after a sigmoid activation, and $T \in \{0,1\}^{B \times 1 \times H \times W}$ denote the ground-truth binary change mask. The weighted BCE term is defined as:

\begin{equation}
A_{i,j,k} = w_i\, T_{i,j,k} \log P_{i,j,k}
\end{equation}

\begin{equation}
B_{i,j,k} = (1 - T_{i,j,k}) \log(1 - P_{i,j,k})
\end{equation}

\begin{equation}
\mathcal{L}_{\text{BCE}} = -\frac{1}{BHW} \sum_{i,j,k} \left( A_{i,j,k} + B_{i,j,k} \right)
\end{equation}

where the pixel-wise weight $w_i$ emphasizes the rare foreground pixels. In practice, we set $w_i = 9.0$ to reflect a typical foreground ratio of $0.1$, balancing the contribution of background and foreground pixels.

The Dice loss further encourages overlap between predicted and ground-truth masks, mitigating the effect of class imbalance:
\begin{equation}
\mathcal{L}_{\text{Dice}} = 1 - \frac{2 \sum P_{i,j,k} T_{i,j,k} + \epsilon}{\sum P_{i,j,k} + \sum T_{i,j,k} + \epsilon},
\end{equation}
where $\epsilon=1e-6$ prevents division by zero.

The final loss is a simple combination of the two terms:
\begin{equation}
\mathcal{L}_{\text{CD}} = \mathcal{L}_{\text{BCE}} + \mathcal{L}_{\text{Dice}}.
\end{equation}

This composite loss effectively balances pixel-level classification and region-level overlap, improving training stability and performance in scenarios with small, sparse change regions.

\section{Experiment}
\subsection{Datasets and Experimental Setup}

We conducted a systematic evaluation of the proposed method on three publicly available remote sensing change detection benchmarks: \textbf{LEVIR-CD}~\cite{sta}, \textbf{LEVIR-CD+}, and \textbf{SYSU-CD}~\cite{SYSU-CD}. The LEVIR-CD dataset contains 637 pairs of Google Earth multi-temporal images with 0.5\,m resolution and size 1024$\times$1024 pixels, covering building construction and demolition changes, and serves as a widely used benchmark in remote sensing building change detection. LEVIR-CD+ is an enhanced version of LEVIR-CD with increased sample size and more diverse change categories. SYSU-CD consists of approximately 20,000 pairs of high-resolution aerial images (64$\times$64 pixels), covering various changes including building expansion, road extension, and vegetation variation. To accommodate the model input and batch training, each image pair from LEVIR-CD and LEVIR-CD+ was evenly cropped into 64$\times$64 sub-images for training, validation, and testing.

For model and training configuration, we adopt Qwen2.5-VL-7B-Instruct~\cite{bai2025qwen2, qwen} as the backbone model, implemented based on the LLaMA-Factory~\cite{llama} and EasyR1~\cite{easyr1} codebases. The training process consists of two stages: SFT stage and RL stage. In the SFT stage, the model is trained for the predefined number of epochs, while in the RL stage, the model is trained for 30 epochs. All experiments are conducted on eight A100 GPUs, with total training time ranging from approximately 30 to 50 hours. During RL, GRPO is applied as the RL paradigm; during SFT, the model is optimized using the Next Token Prediction(NTP)loss for visual instruction fine-tuning.
\begin{table*}[h]
\centering
\caption{Performance comparison of different change detection models on LEVIR-CD+ dataset.}
\label{tab:levir+}
\vspace{-0.3cm}
\normalsize  
\setlength{\tabcolsep}{12pt}  
\begin{tabular}{lccccc}
\toprule
\textbf{Method} & \textbf{Precision (\%)} & \textbf{Recall (\%)} & \textbf{F1 (\%)} & \textbf{IoU (\%)} & \textbf{OA (\%)} \\
\midrule
FC-Siam-conc~\cite{ASTABM}   & 79.94 & 72.55 & 76.06 & 61.37 & 98.14 \\
CDNet~\cite{CDNet}          & 74.12 & 78.45 & 76.22 & 61.58 & 98.01 \\
STANet~\cite{sta}         & 90.68 & 87.70 & 89.17 & 80.45 & 98.91 \\
L-UNet~\cite{L-UNet}         & 78.59 & 82.25 & 80.38 & 67.19 & 98.36 \\
SNUNet~\cite{SNUNet-CD}         & 80.52 & 79.43 & 80.82 & 68.17 & 98.42 \\
DTCDSCN~\cite{DTCDSCN}        & 83.41 & 79.08 & 81.19 & 68.76 & 98.58 \\
BIT~\cite{BIT}            & 80.61 & 82.84 & 81.71 & 69.07 & 98.49 \\
IFNet~\cite{IFNet}          & 83.77 & 80.32 & 82.29 & 70.97 & 98.73 \\
ChangeCLIP~\cite{ChangeCLIP}     & 88.46 & 83.90 & 86.12 & 75.63 & 98.90 \\
\midrule
\textbf{ViLaCD-R1(Ours)} & \textbf{89.51} & 81.94 & \textbf{87.27} & \textbf{78.96} & 98.61 \\
\bottomrule
\end{tabular}
\vspace{-0.3cm}
\end{table*}

\begin{table*}[h]
\centering
\caption{Performance comparison of different change detection models on SYSU-CD dataset.}
\label{tab:sysu}
\vspace{-0.3cm}
\normalsize  
\setlength{\tabcolsep}{12pt}  
\begin{tabular}{lccccc}
\toprule
\textbf{Method} & \textbf{Precision (\%)} & \textbf{Recall (\%)} & \textbf{F1 (\%)} & \textbf{IoU (\%)} & \textbf{OA (\%)} \\
\midrule
CDNet~\cite{CDNet} & 79.34 & 77.29 & 78.30 & 64.34 & 89.90 \\
ISNet~\cite{ISNet} & 76.41 & 80.27 & 78.29 & 64.44 & 90.01 \\
SNUNet~\cite{SNUNet-CD} & 83.58 & 75.87 & 79.54 & 66.02 & 90.79 \\
L-UNet~\cite{L-UNet} & 81.24 & 78.08 & 79.63 & 66.15 & 90.58 \\
IF-Net~\cite{IFNet} & 80.98 & 79.37 & 80.17 & 66.90 & 90.74 \\
DARNet~\cite{DARNet} & 83.04 & 79.11 & 81.03 & 68.10 & 91.26 \\
ICFNet~\cite{ICFNet} & 83.37 & 78.51 & 80.74 & 68.12 & 91.24 \\
ChangeCLIP~\cite{ChangeCLIP} & 87.16 & 79.80 & 83.32 & 71.41 & 92.46 \\
\midrule
\textbf{ViLaCD-R1 (Ours)} & 85.85 & \textbf{84.76} & \textbf{83.71} & \textbf{74.11} & \textbf{95.65} \\
\bottomrule
\end{tabular}
\end{table*}

\subsection{Results}
\begin{figure}[h]
    \centering
    \includegraphics[width=1.0\linewidth]{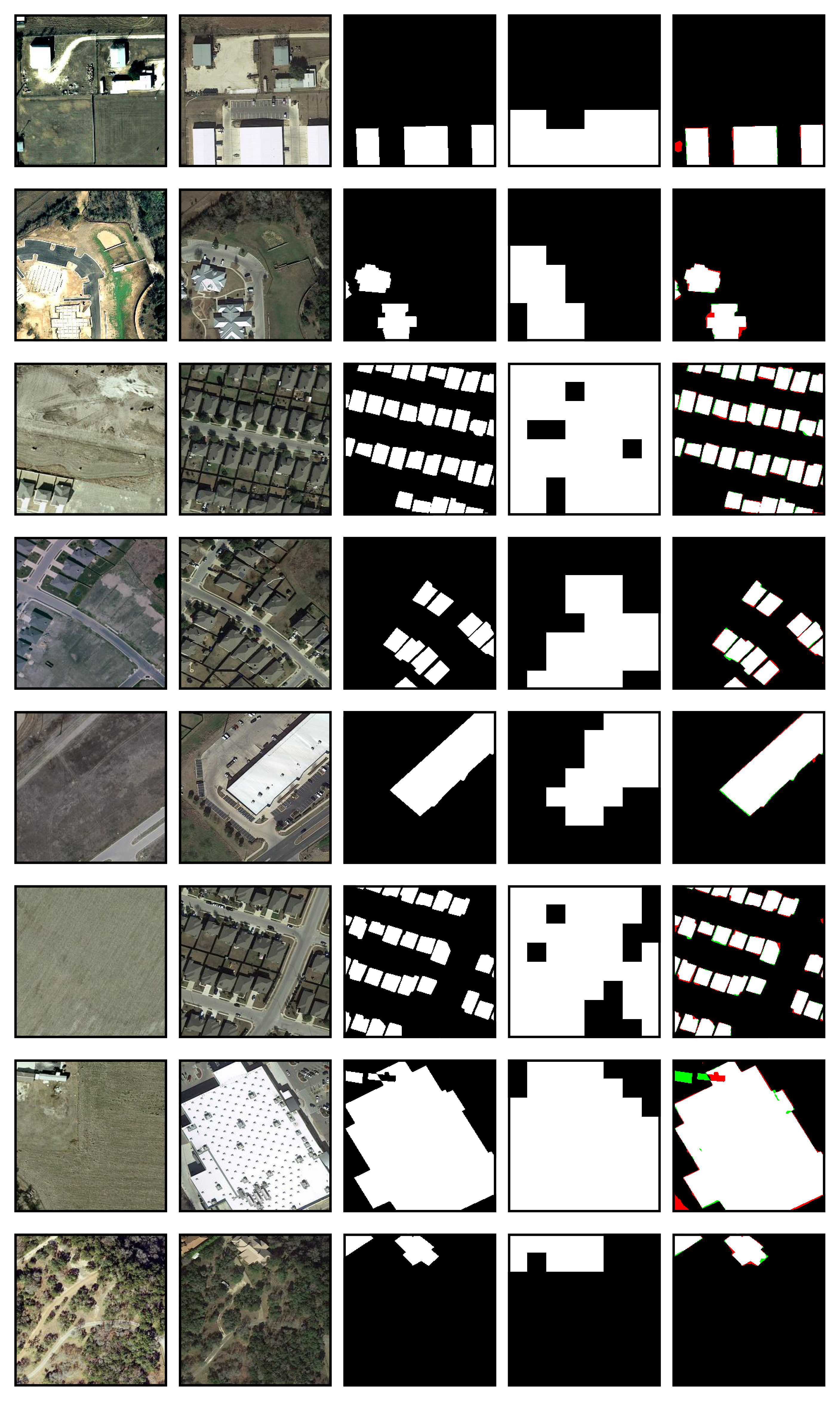}
    \caption{Visulization results of the proposed method on the LEVIR-CD dataset}
    \label{fig:vis}
\end{figure}
We evaluate ViLaCD-R1 using standard metrics in remote sensing change detection, including Intersection over Union (IoU), F1 score, Precision, Recall, and Overall Accuracy (OA). All metrics are computed on the SYSU-CD, LEVIR-CD, and LEVIR-CD+ datasets and compared against state-of-the-art methods. Quantitative results are summarized in Tables~\ref{tab:levir}, \ref{tab:levir+}, and \ref{tab:sysu}.

\noindent\textbf{Main Results.} Experimental results demonstrate that the proposed model significantly outperforms all existing baseline methods across all datasets, achieving the highest IoU metrics. Its exceptional performance on the LEVIR-CD and LEVIR-CD+ datasets validates the model's precise detection capability for small-scale targets and boundary-sensitive change regions. As shown in Fig. \ref{fig:vis}, the visualization results on the LEVIR-CD dataset further indicate that the model maintains accurate identification even when confronted with subtle discrepancies present in the dataset, highlighting ViLaCD-R1's advantage in precise localization of change areas. Moreover, the nearly comprehensive performance improvement achieved on the SYSU-CD dataset proves the model's effectiveness in suppressing false alarms caused by seasonal variations and illumination changes. Collectively, these results fully demonstrate the efficacy of our proposed two-stage hierarchical reasoning framework and mask-guided fine-grained decoder in enhancing semantic comprehension and improving spatial localization accuracy.
\begin{table}[h]
\centering
\caption{Ablation study of ViLaCD-R1 on LEVIR-CD, analyzing the effects of the mask, decoder, and RL modules.}
\label{tab:ablation1}
\footnotesize
\resizebox{\linewidth}{!}{%
\begin{tabular}{lccccc}
\toprule
Ablation & IoU (\%) & F1 (\%) & Pre (\%) & Rec (\%) & OA (\%) \\
\midrule
w/o Mask Guidance & 79.59 & 83.10 & 89.30 & 88.08 & 98.90 \\
w/o GRPO (RL) & 83.20 & 86.96 & 89.76 & 91.37 & 98.76 \\
w/o Decoder & 32.05 & 48.55 & 32.36 & \textbf{97.10} & 89.52 \\
ViLaCD-R1 & \textbf{84.33} & \textbf{87.70} & \textbf{92.76} & 90.84 & \textbf{98.99} \\
\bottomrule
\end{tabular}%
}
\end{table}

\begin{table}[h]
\centering
\caption{Ablation study of ViLaCD-R1 on LEVIR-CD, analyzing the effects of different patch size.}
\label{tab:ablation2}
\footnotesize
\resizebox{\linewidth}{!}{%
\begin{tabular}{lccccc}
\toprule
Ablation & IoU (\%) & F1 (\%) & Pre (\%) & Rec (\%) & OA (\%) \\
\midrule
Patch 4×4 & 83.14 & 86.60 & 91.31 & 89.66 & 98.96 \\
Patch 16×16 & 83.92 & 87.30 & \textbf{94.39} & 88.24 & 98.89 \\
ViLaCD-R1 & \textbf{84.33} & \textbf{87.70} & 92.76 & \textbf{90.84} & \textbf{98.99} \\
\bottomrule
\end{tabular}%
}
\end{table}
\noindent\textbf{Ablation Study.} To evaluate the contribution of each component in ViLaCD-R1, we conducted systematic ablation experiments on the LEVIR-CD dataset by selectively removing or modifying key modules and strategies. The ablation settings include: (i) removing coarse mask guidance, (ii) adjusting the image patch size, and (iii) omitting GRPO-based RL training.

The results, summarized in Table~\ref{tab:ablation1}, show that removing coarse mask guidance significantly reduces F1 and IoU, highlighting its crucial role in guiding fine-grained predictions. Omitting GRPO-based RL training results in an F1 drop of approximately 0.68\% and an IoU drop of about 1.07\%, demonstrating the importance of cross-temporal RL for enhancing semantic understanding and spatial consistency. Table~\ref{tab:ablation2} shows that Adjusting the patch size also affects performance, but to a relatively smaller extent.

Overall, the ablation study confirms that coarse mask guidance, two-stage hierarchical reasoning, and GRPO-based RL training all contribute significantly to the final performance of ViLaCD-R1, indicating that these components collaboratively enhance both semantic comprehension and pixel-level spatial precision.

\subsection{Insightful Analyses}
ViLaCD-R1 achieves accurate remote sensing change detection through a two-stage hierarchical reasoning framework, where SFT provides supervised semantic grounding, coarse mask guidance offers structural priors, and GRPO-based RL further enhances cross-temporal consistency and fine-grained spatial reasoning. Based on the experimental results, we provide the following insightful analyses.

\begin{itemize}
    \item \textit{Effectiveness of Coarse Mask Guidance:}  
    The coarse mask guidance substantially improves spatial precision by supplying an initial localization prior, effectively reducing false alarms and enabling the decoder to refine boundaries with higher reliability.

    \item \textit{Benefits of GRPO-based Temporal Reasoning:}  
    The GRPO-driven RL stage enhances the model’s ability to reason about temporal differences beyond pixel-level variations, explaining the notable performance gains when RL is incorporated.

    \item \textit{Necessity of the Fine-grained Decoder:}  
    The fine-grained decoder is essential for reconstructing accurate change maps. The drastic performance drop when it is removed shows that high-level VLM features alone are insufficient for precise spatial prediction.

    \item \textit{Robustness to Patch-size Variation:}  
    Patch-size experiments demonstrate that ViLaCD-R1 maintains stable performance across different spatial granularities, reflecting the robustness of the hierarchical reasoning and mask-guided refinement mechanisms.
\end{itemize}

\section{Conclusion}

We proposed ViLaCD-R1, a two-stage framework for remote sensing change detection that integrates semantic reasoning with mask-guided fine-grained decoding. By using MIR-generated coarse masks to guide deep feature integration and spatial refinement, ViLaCD-R1 effectively addresses the limitations of both feature-based and VLM-based methods, achieving robust boundary reconstruction and accurate detection of small and subtle changes. Experiments on SYSU-CD, LEVIR-CD, and LEVIR-CD+ show consistent improvements across IoU, F1, precision, recall, and overall accuracy, while ablation studies confirm the synergistic benefits of coarse mask guidance, hierarchical reasoning, patch-level modeling, and RL.

Although ViLaCD-R1 achieves strong performance, it still relies on the stability of coarse mask generation and currently focuses on binary change detection. Future work will explore extending the framework to multi-class scenarios, improving robustness through temporal attention or self-supervised pretraining, and incorporating lightweight uncertainty estimation mechanisms to enhance interpretability and ensure reliability in real-world applications.

{
    \small
    \bibliographystyle{ieeenat_fullname}
    \bibliography{main}
}

\end{document}